\documentclass[conference]{IEEEtran}
\IEEEoverridecommandlockouts

\usepackage{cite}
\usepackage{amsmath,amssymb,amsfonts}
\usepackage{algorithmic}
\usepackage{graphicx}
\usepackage{marvosym} 
\usepackage{amssymb}
\usepackage{amsmath}
\usepackage{stfloats}
\usepackage{bbm}
\usepackage{url}
\usepackage{amssymb}
\usepackage{amsmath,epsfig}
\usepackage{balance}
\usepackage{booktabs}
\usepackage{graphicx}
\usepackage{cite}
\usepackage{subfigure}
\usepackage{multirow}
\usepackage{hyperref}
\usepackage{textcomp}
\usepackage{xcolor}
\def\BibTeX{{\rm B\kern-.05em{\sc i\kern-.025em b}\kern-.08em
    T\kern-.1667em\lower.7ex\hbox{E}\kern-.125emX}}

\title{CoMT: Chain-of-Medical-Thought Reduces Hallucination in Medical Report Generation\\
\thanks{$^{\dag}$Co-first authors. $^{\textrm{\Letter}}$Corresponding authors.}
}

\author{\IEEEauthorblockN{Yue Jiang\textsuperscript{1,4 \dag}, 
Jiawei Chen\textsuperscript{1,4 \dag}, 
Dingkang Yang\textsuperscript{1,4 \Letter}, 
Mingcheng Li\textsuperscript{1,4}, \\
Shunli Wang\textsuperscript{1, 2, 4},
Tong Wu\textsuperscript{2,3}, 
Ke Li\textsuperscript{2}, 
Lihua Zhang\textsuperscript{1,4,5,6 \Letter}}
\IEEEauthorblockA{\textsuperscript{1} Academy for Engineering and Technology, Fudan University}
\IEEEauthorblockA{\textsuperscript{2} Tencent YouTu Lab}
\IEEEauthorblockA{\textsuperscript{3} Xiamen University}
\IEEEauthorblockA{\textsuperscript{4} Cognition and Intelligent Technology Laboratory (CIT Lab), Institute of Meta-Medical, Fudan University}
\IEEEauthorblockA{\textsuperscript{5} Engineering Research Center of AI and Robotics, Ministry of Education, Shanghai, China}
\IEEEauthorblockA{\textsuperscript{6} Jilin Provincial Key Laboratory of Intelligence Science and Engineering, Changchun, China}
\vspace{-30pt}
}

\begin{document}

\maketitle

\begin{abstract}
Automatic medical report generation (MRG), which possesses significant research value as it can aid radiologists in clinical diagnosis and report composition, has garnered increasing attention. Despite recent progress, generating accurate reports remains arduous due to the requirement for precise clinical comprehension and disease diagnosis inference. Furthermore, owing to the limited accessibility of medical data and the imbalanced distribution of diseases, the underrepresentation of rare diseases in training data makes large-scale medical visual language models prone to hallucinations, such as omissions or fabrications, severely undermining diagnostic performance and further intensifying the challenges for MRG in practice. In this study, to effectively mitigate hallucinations in medical report generation, we propose a chain-of-medical-thought approach (CoMT), which intends to imitate the cognitive process of human doctors by decomposing diagnostic procedures. The radiological features with different importance are structured into fine-grained medical thought chains to enhance the inferential ability during diagnosis, thereby alleviating hallucination problems and enhancing the diagnostic accuracy of MRG. The code and dataset have been released at \url{https://github.com/FRENKIE-CHIANG/CoMT}.
\end{abstract}

\begin{IEEEkeywords}
 medical report generation, hallucination, chain-of-thought.
\end{IEEEkeywords}

\vspace{-6pt}

\section{Introduction}
\vspace{-2pt}
Radiological images and the corresponding reports are extensively utilized in clinical diagnosis~\cite{Delrue_Gosselin_Ilsen_Landeghem_Duyck_Mey_2011}. However, the composition of medical reports demands substantial specialized domain knowledge, as only seasoned radiologists can conduct a thorough analysis of radiological images and accurately infer diseases.  Medical report generation (MRG) aims to automatically generate free-text descriptions of radiological images, thereby alleviating the burden on radiologists and facilitating auxiliary diagnosis.  Nevertheless, generating precise medical reports is challenging, as it requires comprehensive analysis to ensure that any detailed features potentially indicating underlying diseases are captured, as any omission would lead to severe consequences.
In recent years, a multitude of studies on MRG have emerged.  With the advancement of large-scale visual language models (LVLMs), the clinical applicability of MRG has attracted renewed attention, and diverse methods have been proposed to enhance the performance. For instance, multi-task learning~\cite{Jing_Xie_Xing_2018, Yan_Pei_2022} has been prevalently employed to obtain superior feature representations by undertaking additional auxiliary tasks, such as disease classification~\cite{Jin_Che_Lin_Chen_2024}, thereby improving the diagnostic capabilities of the model for radiological images.  Additionally, contrastive learning~\cite{Liu_Yin_Wang_Ge_Zhang_Sun_2021} is another efficacious technique that enhances feature learning and diagnostic ability by contrasting abnormal radiological images with normal ones to locate disease regions.
Despite these achievements, existing research has inadequately addressed the adverse impact of hallucinations in LVLM on MRG performance.  When confronted with complex and comprehensive medical issues, LVLM frequently produces responses that seem accurate and fluent but are fundamentally incorrect, which impedes their practical application in real-world medical scenarios.

The reports generated from LVLM may contain various types of hallucinations. Primarily, radiological images may exhibit a wide range of diseases, but the limited availability of medical data, coupled with strict privacy policies and ethical constraints, exacerbates the imbalance in disease representation. The underrepresentation of rare diseases in the training data makes it difficult for the model to learn precise feature representations for various types of diseases, leading to hallucinations such as omissions or fabrications.
In addition, radiological datasets are usually dominated by normal radiological images, resulting in an uneven distribution of data compared to the fewer abnormal images. Within individual images, normal regions often outnumber abnormal ones. This data bias prevents the model from capturing rare but critical abnormal regions like lesion regions. As a result, the model is prone to generating reports that appear plausible but fail to explicitly mention abnormalities, leading to significant omissions in the reports.
Furthermore, in medical reports, descriptions related to local attributes in radiology - such as shape, size and location - typically only appear when abnormalities are present in the images. For dominant normal images, attribute descriptions are often missing, which hinders the model from learning about the boundary between normal and abnormal radiological attributes, resulting in attribute-related hallucinations.

\begin{figure*}[t]
    \centering
    \includegraphics[width=\linewidth]{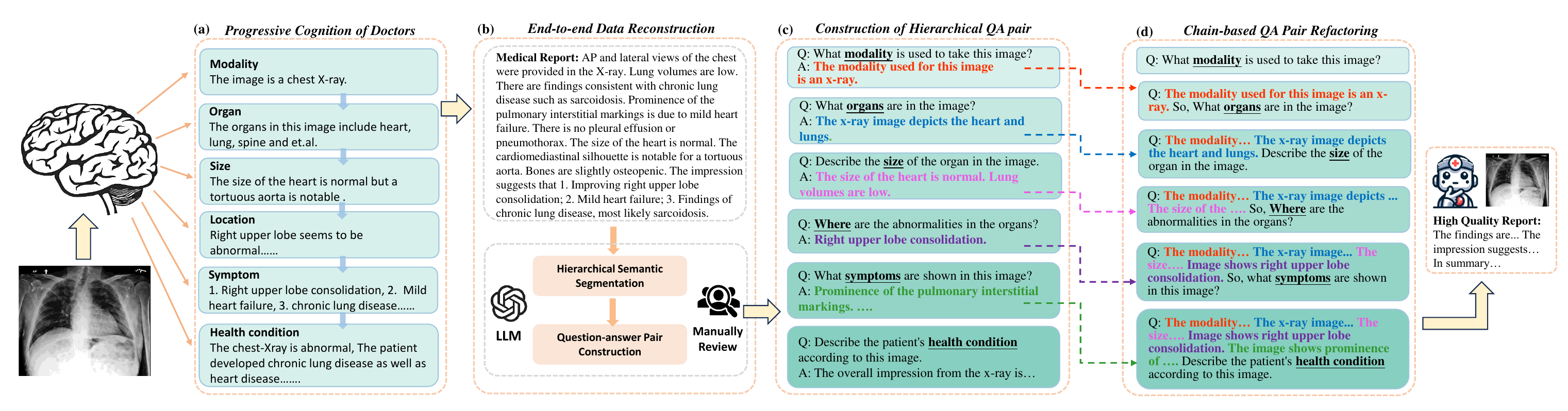}
    \caption{Illustration of CoMT's process for constructing hierarchical QA pairs based on real clinical image reports. (a). The progressive cognitive process of doctors for a medical image. (b). End-to-end data reconstruction of medical reports. (c). Construction of Hierarchical QA pairs. (d). Chain-based QA Pair Refactoring.}
    \vspace{-15pt}
    \label{fig1}
\end{figure*}

To address the aforementioned challenges, we propose a novel method called chain-of-medical-thought (\textbf{CoMT}), which simulates the cognitive steps of human understanding and the diagnostic process of radiologists. CoMT decomposes the unstructured medical reports into fine-grained clues and then links them into a chain of diagnostic thought. CoMT enables the model to learn how to reason about diseases based on details and how to make inductive descriptions rely on local features, thereby reducing various types of hallucinations and improving the reliability of the generated reports.
We evaluated our method on two widely-used benchmarks. The proposed approach outperformed baselines in natural language generation (NLG) metrics, hallucination metrics, and human doctor evaluation, demonstrating the effectiveness of our method.

In summary, our contributions are as follows::
\begin{itemize}
    \item We propose a novel method, CoMT, which decomposes and reconstructs medical reports into chains of medical thought, enhancing the model's diagnostic reasoning ability on MRG. Our method can be easily integrated into existing baselines to improve their performance.
    \item We pioneeringly focus on the hallucination in MRG and reduce the various types of hallucinations commonly found in report generation by CoMT, thereby improving the clinical reliability of MRG.
    \item We demonstrate the superiority of CoMT on two popular MRG benchmarks. And we proposed two new MRG datasets based on CoMT, which will be published soon.
\end{itemize}


\section{Related Work}
\vspace{-2pt}
Recent studies have introduced various methods to enhance the performance of MRG. However, current visual-text generation techniques often generate sentences that resemble natural language but fail to effectively align with visual features. While some approaches~\cite{Liang_Hu_Zhang_Gan_Xing_2017, Anderson_He_Buehler_Teney_Johnson_Gould_Zhang_2017} attempt to address the issue, most neglect the inherent knowledge structure within medical report data. Knowledge graph~\cite{Zhang_Wang_Xu_Yu_Yuille_Xu_2020, Liu_Wu_Ge_Fan_Zou_2021}, which integrates domain knowledge into the model, is an effective technique for improving feature learning and diagnostic capabilities, thereby assisting in report generation. Another widely used technique in MRG to enhance representation learning is multi-task learning. Some approaches~\cite{Jing_Xie_Xing_2018, Wang_Han_Wang_Zhou, Yan_Pei_2022} utilize disease classification to learn discriminative features better.~\cite{Jin_Che_Lin_Chen_2024} makes use of the diagnostic results from the classification via prompts to explicitly guide the generation process. Additionally,~\cite{Liu_Yin_Wang_Ge_Zhang_Sun_2021} is based on contrast learning, using contrastive attention to focus on abnormal regions of images, alleviating the problem of uneven data distribution. But to the best of our knowledge, none of these existing studies have focused on hallucinations in MRG tasks.

One approach to mitigating hallucinations is by increasing the quality as well as the diversity of the training data~\cite{chen2023mitigating, wang2024mitigating} to increase the robustness and generalization of the model. However, traditional data augmentation methods such as image flipping, rotation, scaling, and the introduction of synthetic noise are not suitable for the multimodal medical domain due to the high fidelity required in medical image interpretation. Any alteration that distorts the medical reality of the image or disrupts the feature alignment between images and text can lead to misdiagnoses or the overlooking of critical patient-specific details.
Expanding data by simply rewriting or rephrasing sentences with large language models (LLMs) has received some favour~\cite{wang2024mitigating, rewriting, miss}, but the increase in data diversity from it is constrained.

The chain of thought (CoT)~\cite{2022Chain} is a popular technique used in LLM, that involves generating a series of step-by-step reasoning processes. CoT allows the model to unveil each step of its reasoning, enabling a deeper understanding of the problem and reducing errors caused by abrupt or incomplete thinking. Consequently, it helps models to better address complex problems. Some studies~\cite{chen2024codinterpretablemedicalagent, Shi_Manda_Chowdhury_Arun_Zhu_Lam_2024} have applied CoT to improve the performance of medical models, demonstrating the effectiveness in medical field.
\vspace{-3pt}

\section{Methods}
\vspace{-3pt}
\label{method}
\subsection{Overview of the CoMT}
\vspace{-2pt}
Medical texts authored by humans are often disorganized, complicating the process of correlating visual features with unstructured free texts. Our proposed method \textbf{CoMT} is rooted in a chain-like hierarchy of medical attribute importance. This is inspired by the radiologists' cognitive steps during diagnosis, involving the sequential recognition and integration of information across various levels. In this way, we incorporated the paradigm of logical reasoning in radiology into training, aiming to enable the model to simulate the diagnosis reasoning capabilities of radiologists as closely as possible.
\vspace{-4pt}

\subsection{Inspiration for the CoMT}
\vspace{-3pt}
Exploring whether LVLMs can imitate human cognitive processes~\cite{mischler2024contextual} to infer with fewer hallucinations is a question worth considering. Research in cognitive neuroscience indicates that humans typically process graphical information from superficial to deep levels~\cite{clarke2015understanding}, seamlessly understanding hierarchical information. However, LVLMs lack this capability and must analyze each image patch step by step by reducing the receptive field~\cite{radford2021learningtransferablevisualmodels}. Broadly speaking, the integration of textual information can assist AI models in emulating human cognitive processes.

Figure~\ref{fig1}(a) illustrates the cognitive process of a doctor assessing a medical image. The process typically begins with global information about the image, such as the modality, and progressively focuses on specific organs, followed by fine-grained attributes such as the shape or location of organs or abnormalities. After thoroughly understanding this information, the human brain synthesizes the symptoms presented in the image, ultimately producing an unstructured image report. Therefore, we introduced this thought process into model training through CoMT, allowing LVLMs to mimic human thought steps to generate more accurate medical reports.
\vspace{-3pt}

\subsection{Application Details of CoMT}
\vspace{-3pt}

\label{2.4}
\textbf{Construction of Hierarchical QA pairs: } To transform disorganized reports into high-quality QA pairs that mimic human cognition, we utilize powerful LLMs to semantically segment the image reports. As depicted in Figure~\ref{fig1}(b), the image reports are processed by GPT-4~\cite{gpt-4v} for end-to-end semantic segmentation, and they are broken down into six hierarchical dimensions: \textit{modality}, \textit{organ}, \textit{size}, \textit{abnormal location}, \textit{symptoms}, and \textit{overall health condition}. This segmentation results in six distinct questions, as shown in Figure~\ref{fig1}(c), which independently query and answer the information from different dimensions of the image based on the original reports.

To prevent any potential biases that might be introduced during segmentation by GPT-4 and propagated along the chain of medical thought, we arranged for three researchers from the medical field to conduct two rounds of inspection and verification of the segmentation results.

\textbf{Chain-based QA Pair Refactoring:} To further emulate human cognitive processes, we propose a chain-based QA pair refactoring method, illustrated in Figure~\ref{fig1}(d). For each original question, all previous lower-level answers are used as a prelude to the higher-level question, thereby combining with the original question to reconstruct each QA pair in a chain-like manner.
\vspace{-4pt}
\subsection{Evaluation Metrics}
\vspace{-3pt}
\textbf{NLG Metrics:}
Primarily we perform the automatic evaluation to conduct a fair comparison, using NLG metrics which include BERTScore~\cite{2019BERTScore}, METEOR~\cite{2005METEOR}, and ROUGE-1/2/L~\cite{Lin_2004}.

\textbf{Hallucination Metric:}
To verify the effectiveness of CoMT in mitigating hallucinations, we employed the medical hallucination metric, MediHall Score~\cite{chen2024detecting}. It defines five common types of hallucinations specific to medical texts, three of which are suitable for MRG: catastrophic hallucination (omissions or fabrications of diseases), critical hallucination (misjudging the type of diseases), and attribute hallucination (misjudging attributes such as shape, size, location). The MediHall Score is calculated as a weighted sum based on different types of hallucinations, and the calculation formula is given as:
\vspace{-5pt}
\[
\text{MediHall score} = \frac{1}{N} \sum_{i=1}^{N} S_i ,
\]
\vspace{-10pt}

where \( N \) is the total number of sentences in a report, and \( S_i \) is the score of the \( i \)-th sentence based on its hallucination type. Specifically, the scores of catastrophic, critical, and attribute hallucination are 0, 0.3, and 0.6 respectively, and the score of the correct statement (without hallucination) is 1. Each sentence is judged by GPT-4 and Gemini~\cite{geminiteam2024geminifamilyhighlycapable} for the type of hallucination, and if they differ, three experienced doctors give the final judgment. So,  \( S_i \) is given as:

\vspace{-5pt}
\[
S_i = 
\begin{cases} 
\text{LLM Score}, & \text{if GPT-4 and Gemini} \\
                 & \text{reach the same judgment} \\
\text{Human Score}, & \text{otherwise}
\end{cases}
\]
\vspace{-5pt}

\textbf{Human Doctor Evaluation:}
To align the radiologists' tendency, we followed~\cite{Liu_Yin_Wang_Ge_Zhang_Sun_2021} and conducted human doctor evaluations. 
We invited two professional clinicians to independently evaluate the results, including faithfulness, comprehensiveness, and fluency. The evaluation score from each clinician is calculated by the formula:
\vspace{-6pt}

\[
\text{Human score}=\left( \frac{Num\_Faith + Num\_Com + Num\_Flu}{3 * Num\_Data} \right),
\]
\vspace{-6pt}

where \(Num\_Faith\), \(Num\_Com\), and \(Num\_Flu\) represent the number of faithful, comprehensive, and fluent results, respectively. \(Num\_Data\) represents the size of the dataset. 
\vspace{-10pt}

\section{Experiment and Discussion}
\vspace{-3pt}

\subsection{New Datasets Construction}
\vspace{-3pt}

Following the aforementioned steps in section~\ref{2.4}, based on the MIMIC-CXR dataset~\cite{johnson2019mimiccxrjpg} and the OpenI~\cite{openi} dataset——both of which comprise a vast collection of radiological images accompanied by medical reports——we process these reports following~\cite{chen2024detecting} and construct two new CoMT datasets composed of the chain-based QA pairs (as shown in Figure~\ref{fig1}(d)). Additionally, in order to verify that the effectiveness of CoMT comes from the cognitive process simulation, rather than the increase of data diversity or quantity, we use GPT-4 to rephrase original medical reports, referring to~\cite{rewriting}, so as to expand data diversity and quantity.

\subsection{Effectiveness and Superiority of CoMT}
To verify the effectiveness of CoMT, we compare the performance of models fine-tuned using three types of data: the original MRG medical reports ($\spadesuit$), the data simply rephrased by GPT-4 ($\clubsuit$), and the data constructed by CoMT.
Table~\ref{tab1} and Table~\ref{tab2} present the results on two benchmarks OpenI and MIMIC-CXR.
As shown in the results, models trained using the CoMT data not only demonstrated improvements in NLG metrics but also surpassed the models trained with the original MRG data by approximately 2\% to 5\% in the hallucination metric. This indicates that our CoMT effectively enhances the model's understanding of the global information in images and reduces hallucinations in MRG. Moreover, our CoMT outperformed the models trained on GPT-4 rephrasing data with approximately 5\% to 8\% margin in terms of MediHall, which proves that the improved performance is due to the introduction of the chain of medical diagnostic thought, rather than the increase of data diversity or quantity.
\vspace{-6pt}
\begin{table}[t]
\setlength{\tabcolsep}{5pt}
\centering
\caption{Comparison of the different methods on OpenI.\quad ``R-1/2/L'' means the ROUGE-1/2/L. ``BS'', ``MT'' stand for the BERTScore and METEOR.\quad ``$\spadesuit$'', ``$\clubsuit$'' stand for the original MRG data and the GPT-4 rephrasing data.}
\vspace{-3pt}
\resizebox{0.8\linewidth}{!}{%
\begin{tabular}{cccccccc}
\hline
Model                    & BS        & MT       & R-1     & R-2     & R-L       & MediHall & Human \\ \hline
LLaVA-Med~\cite{llavamed} + $\spadesuit$         & 57.87         & 18.18        & 22.81       & 4.01       & 21.29           & 0.393     &   0.227       \\
LLaVA-Med + $\clubsuit$  & 57.6         & 18.34        & 22.88       & 3.94       & 21.37            & 0.379       & 0.207       \\
\textbf{LLaVA-Med + CoMT}       & \textbf{66.21}         & \textbf{18.96}        & \textbf{31.57}       & \textbf{8.79}       & \textbf{31.10}     & \textbf{0.427}          &   \textbf{0.239 }\\ \hline
MiniGPT4~\cite{minigpt} + $\spadesuit$          & 60.18         & 19.82        & 24.89       &5.12        & 25.12          & 0.620          &   0.594 \\
MiniGPT4 + $\clubsuit$   & 59.41         & 20.10        & 24.06       & 4.97       & 21.30          & 0.603       &   0.620  \\
\textbf{MiniGPT4 + CoMT }       & \textbf{61.94} & \textbf{21.71}  & 24.75    & \textbf{5.96}  & \textbf{27.06}    & \textbf{0.663}   & \textbf{0.651}\\ \hline
XrayGPT~\cite{xraygpt} + $\spadesuit$           & 64.71         & 20.97        & 30.24       & 6.97       & 27.58            & 0.641        &   0.616   \\
XrayGPT + $\clubsuit$    & 64.69         & 21.30       & 30.22       & 6.59       & 27.65            & 0.613          &  0.587  \\
\textbf{XrayGPT + CoMT}         & \textbf{65.57}         & \textbf{22.03}        & \textbf{31.14}       & \textbf{7.57}       & \textbf{27.92}       & \textbf{0.718 }         &   \textbf{0.622} \\ \hline
mPLUG-Owl2~\cite{ye2023mplug} + $\spadesuit$            & 69.77         & 35.38         & 40.39          & 12.91          & 36.67                & 0.688  &  0.498 \\
mPLUG-Owl2 + $\clubsuit$     & 61.72         & 35.86         & 40.10           & 12.31          & 36.60                   & 0.654 &   0.513 \\
\textbf{mPLUG-Owl2 + CoMT} & \textbf{70.89} & \textbf{37.57} & \textbf{42.06} & \textbf{13.23} & \textbf{36.89}  & \textbf{0.736} &  \textbf{0.574} \\ \hline
R2Gen~\cite{chen-emnlp-2020-r2gen} + $\spadesuit$   & 57.99 & 15.22 & 20.90 & 3.82  & 25.14 & 0.474    & 0.182\\
R2Gen + $\clubsuit$      & 57.63 & 15.31 & 20.59 & 3.87  & 26.05 & 0.477    & 0.193\\
\textbf{R2Gen + CoMT }   & \textbf{60.10} & \textbf{15.97} & 20.35 & \textbf{4.26}  & \textbf{26.62} & \textbf{0.510 }   & \textbf{0.195}\\ \hline
\end{tabular}
}
\label{tab1}
\vspace{-9pt}
\end{table}

\begin{table}[t]
\setlength{\tabcolsep}{5pt}
\centering
\caption{Comparison of the different methods on MIMIC-CXR.}
\vspace{-3pt}
\resizebox{0.8\linewidth}{!}{%
\begin{tabular}{cccccccc}
\hline
Model                    & BS        & MT       & R-1     & R-2     & R-L       & MediHall & Human \\ \hline
LLaVA-Med + $\spadesuit$ & 56.11 & 13.04 & 20.94 & 3.12  & 23.21 & 0.371    & 0.188  \\
LLaVA-Med + $\clubsuit$  & 55.28 & 14.33 & 20.34 & 3.09  & 24.77 & 0.392    & 0.195    \\
\textbf{LLaVA-Med + CoMT}&\textbf{ 60.02} & \textbf{17.87} & 20.67 & \textbf{5.11}  & \textbf{26.76} &\textbf{ 0.449 }   & \textbf{0.197 }  \\ \hline
MiniGPT4 + $\spadesuit$  & 62.84 & 19.19 & 23.46 & 4.36  & 25.42 & 0.604    & 0.424  \\
MiniGPT4 + $\clubsuit$   & 60.59 & 24.21 & 24.71 & 4.40  & 26.10 & 0.629    & 0.448   \\
\textbf{MiniGPT4 + CoMT }& \textbf{63.07 }& 23.90 & \textbf{25.55} &\textbf{ 4.82}  & \textbf{26.19} &\textbf{ 0.658 }   &\textbf{ 0.463} \\ \hline
XrayGPT + $\spadesuit$   & 66.30 & 25.76 & 24.90 & 7.03  & 26.33 & 0.676    & 0.457      \\
XrayGPT + $\clubsuit$    & 66.52 & 24.31 & 24.91 & 6.87  & 27.31 & 0.660    & 0.470    \\
\textbf{XrayGPT + CoMT}  &\textbf{ 66.93 }&\textbf{ 27.28} &\textbf{ 26.10} & \textbf{6.95}  &\textbf{ 27.89} & \textbf{0.695  }  & \textbf{0.492} \\ \hline
mPLUG-Owl2 + $\spadesuit$& 72.35 & 29.42 & 29.89 & 13.37 & 26.93 & 0.640    & 0.338   \\
mPLUG-Owl2 + $\clubsuit$ & 68.47 & 29.33 & 28.74 & 13.19 & 27.02 & 0.629    & 0.325   \\
\textbf{mPLUG-Owl2 + CoMT}& \textbf{73.09} & \textbf{31.19 }&\textbf{ 30.03} &\textbf{ 13.46} &\textbf{ 28.31} &\textbf{ 0.681  }  &\textbf{ 0.369 } \\ \hline
R2Gen + $\spadesuit$       & 59.31 & 14.23 & 22.71 & 4.01  & 27.70 & 0.492    & 0.209 \\
R2Gen + $\clubsuit$      & 58.43 & 14.99 & 21.93 & 3.99  & 27.21 & 0.515    & 0.196 \\
\textbf{R2Gen + CoMT }   & \textbf{59.40} & \textbf{16.52 }& 22.49 &\textbf{ 5.37 } &\textbf{ 29.33 }&\textbf{ 0.548}    & \textbf{0.237 }\\ \hline
\end{tabular}
}
\label{tab2}
\vspace{-13pt}
\end{table}

\begin{figure}[t]
    \centering
    \vspace{-10pt}
    \includegraphics[width=0.8\linewidth]{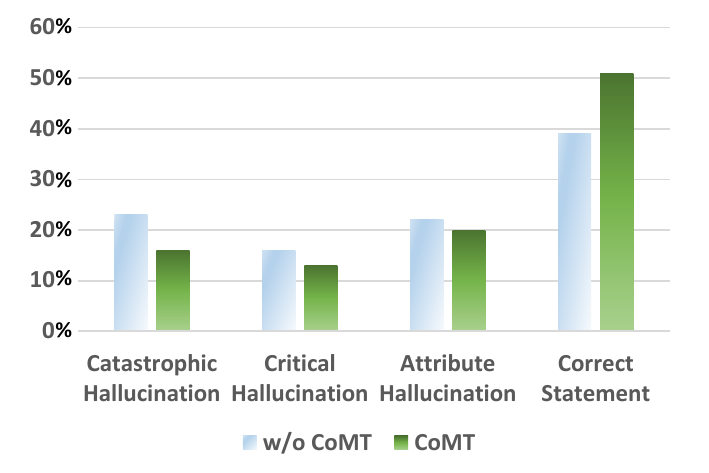}
    \vspace{-10pt}
    \caption{The percentage change in the number of different types of hallucinations after using the CoMT method. ``w/o'' stands for without.}
    \label{fig2}
\end{figure}

\vspace{-3pt}

\begin{table}[h]
\centering
\caption{Comparison of CoMT and traditional Data Augmentation methods for Medihall Score on OpenI.  ``w/o'' stands for without data augmentation. ``GN'' stands for Gaussian noise.} 
\resizebox{0.8\linewidth}{!}{%
\begin{tabular}{ccccccccc}
\hline
Model       & w/o       & Insert & Swap   & Delete & GN   & Flip   & Clip & CoMT  \\ \hline
LLaVA-med   & 0.393     & 0.402 & 0.399 & 0.402 & 0.379 & 0.389 & 0.396 & \textbf{0.427} \\
MiniGPT4    & 0.620     & 0.645 & 0.621 & 0.632 & 0.601  & 0.597 & 0.615 & \textbf{0.663} \\ 
XrayGPT     & 0.641     & 0.666 & 0.628 & 0.552 & 0.560 & 0.599 & 0.620 & \textbf{0.718} \\
mPLUG-Owl2  & 0.688     & 0.694  & 0.682 & 0.687 & 0.682 & 0.684  & 0.681 & \textbf{0.736}\\ 
R2Gen       & 0.489     & 0.472  & 0.481 & 0.482 & 0.479 & 0.4807  & 0.495 & \textbf{0.510}\\ \hline
\end{tabular}
}
\label{tab3}
\vspace{-10pt}
\end{table}

\begin{table}[ht]
\centering
\vspace{-10pt}
\caption{Out-of-Distribution (OOD) capacities of different models with different augmentation methods. ``$\varnothing$'' and ``$\spadesuit$'' stand for no fine-tuning and fine-tuning with original medical reports.} 
\resizebox{0.8\linewidth}{!}{%
\begin{tabular}{ccccccc}
\hline
Model               & BERTSore     & METEOR          & ROUGE-L        & MediHall & Human \\ \hline
LLaVA-Med + $\varnothing$           & 50.94 & 10.92  & 13.72   & 0.292  & 0.161\\
LLaVA-Med + $\spadesuit$    & 51.40  & 11.04  & 13.91  & 0.275  & 0.189\\
\textbf{LLaVA-Med + CoMT}  & \textbf{58.11} & \textbf{16.39}  & \textbf{16.88}  & \textbf{0.341} &  \textbf{0.207}\\ \hline
R2Gen + $\varnothing$           & 53.01 & 9.92  & 10.90   & 0.302  & 0.159\\
R2Gen + $\spadesuit$    & 52.29  & 9.04  & 11.24  & 0.315  & 0.161\\
\textbf{R2Gen + CoMT}  & \textbf{55.17} & \textbf{10.33}  & \textbf{12.37}  & \textbf{0.358} &  \textbf{0.170}\\ \hline
MiniGPT4 + $\varnothing$            & 56.24 & 14.20 & 12.63  & 0.329 &  0.338\\
MiniGPT4 + $\spadesuit$     & 60.54 & 23.09  & 19.21  & 0.411 &  0.362\\
\textbf{MiniGPT4 + CoMT}  & 60.12 & 21.90   & \textbf{19.94 }  & \textbf{0.452 } & \textbf{0.391}\\ \hline
mPLUG-Owl2 + $\varnothing$          & 55.01 & 9.91 & 13.01& 0.350 &  0.273\\
mPLUG-Owl2 + $\spadesuit$   & 55.16 & 9.60   & 13.23  & 0.363 &  0.295\\
\textbf{mPLUG-Owl2 + CoMT} & \textbf{55.48} & \textbf{10.01}  & \textbf{13.49}  & \textbf{0.416} &\textbf{ 0.319} \\ \hline

\end{tabular}
}
\label{tab4}
\vspace{-10pt}
\end{table}

Additionally, as shown in Table~\ref{tab3}, compared CoMT with traditional data augmentation methods for text and images, the traditional methods expand data quantity and enhance the model's robustness, but they do not result in a significant improvement in the MediHall score. In contrast, CoMT shows the superiority in MediHall Score, which further demonstrates that emulating human cognitive processes by CoMT can directly reduce the hallucinations in LVLMs for MRG.

\subsection{Generalizability of CoMT on Out-of-Distribution Dataset}
\vspace{-4pt}
To verify whether CoMT can enhance the model's generalization capability on the out-of-distribution (OOD) dataset, we fine-tune baseline models on the OpenI and evaluate the performance on the MIMIC-CXR.
As shown in Table~\ref{tab4}, models fine-tuned by CoMT data significantly outperform other methods in most metrics. Especially there are approximately 6\% to 12\% increase and 5\% increase in hallucination metric and human doctor evaluation respectively.

\subsection{Effects of CoMT on Different Types of Hallucinations}
As shown in Figure~\ref{fig2}, the CoMT method reduced the number of different types of hallucinations on average, with the most significant reduction in catastrophic hallucinations and a prominent increase in correct statements. This suggests the introduction of CoMT helps models to follow the details in the original medical reports for analysis, thus reducing hallucinations, especially the more serious types, such as omissions and fabrications.

\section{Conclusions}
\vspace{-3pt}
In this paper, we proposed a method called CoMT, that mimics human cognitive processes to tackle the problem of hallucination in MRG. CoMT leverages the cognitive processes of human doctors by segmenting medical text into different attribute layers and constructing chain-based QA pairs. Experiments on two datasets demonstrate the superiority of our method, particularly its advantage in generating accurate and comprehensive reports with fewer hallucinations, bridging the gap between current MRG methods and the clinical demands in real-world medical scenarios.

\section{Acknowledgements}
This work is supported by the Shanghai Municipal Science and Technology Major Project (2021SHZDZX0103).

\end{document}